\documentclass[journal]{IEEEtran}
\usepackage{lineno,hyperref}
\usepackage{bbm}
\usepackage{bm}
\usepackage{amsfonts}
\usepackage{amsmath}
\usepackage{mathrsfs}
\usepackage{amssymb}
\usepackage{enumerate}
\usepackage{graphicx}
\usepackage{subfigure}
\usepackage{booktabs}
\usepackage{graphicx}
\usepackage{setspace}
\usepackage{algorithm}
\usepackage{algorithmic}
\usepackage{setspace}
\usepackage{multirow}
\usepackage{array}
\usepackage{color}
\usepackage{cite}
\usepackage{geometry}
\ifCLASSINFOpdf
\else
\fi
\begin{document}
\title{Image-Specific Information Suppression and Implicit Local Alignment for Text-based Person Search}
\author{Shuanglin~Yan, Hao~Tang, Liyan~Zhang, Jinhui~Tang, \emph{Senior Member, IEEE}
\thanks{S. Yan, H. Tang, and J. Tang are with the School of Computer Science and Engineering, Nanjing University of Science and Technology, Nanjing 210094, China  (e-mail: shuanglinyan@njust.edu.cn; tanghao0918@njust.edu.cn; jinhuitang@njust.edu.cn).}
\thanks{L. Zhang is with the College of Computer Science and Technology, Nanjing University of Aeronautics and Astronautics, Nanjing 210016, China (e-mail: zhangliyan@nuaa.edu.cn).}
}
\markboth{Journal of \LaTeX\ Class Files}%
{Shell \MakeLowercase{\textit{et al.}}: Bare Demo of IEEEtran.cls for IEEE Journals}
\maketitle
\begin{abstract}
Text-based person search (TBPS) is a challenging task that aims to search pedestrian images with the same identity from an image gallery given a query text. In recent years, TBPS has made remarkable progress and state-of-the-art methods achieve superior performance by learning local fine-grained correspondence between images and texts. However, most existing methods rely on explicitly generated local parts to model fine-grained correspondence between modalities, which is unreliable due to the lack of contextual information or the potential introduction of noise. Moreover, existing methods seldom consider the information inequality problem between modalities caused by image-specific information. To address these limitations, we propose an efficient joint Multi-level Alignment Network (MANet) for TBPS, which can learn aligned image/text feature representations between modalities at multiple levels, and realize fast and effective person search. Specifically, we first design an image-specific information suppression module, which suppresses image background and environmental factors by relation-guided localization and channel attention filtration respectively. This module effectively alleviates the information inequality problem and realizes the alignment of information volume between images and texts. Secondly, we propose an implicit local alignment module to adaptively aggregate all pixel/word features of image/text to a set of modality-shared semantic topic centers and implicitly learn the local fine-grained correspondence between modalities without additional supervision and cross-modal interactions. And a global alignment is introduced as a supplement to the local perspective. The cooperation of global and local alignment modules enables better semantic alignment between modalities. Extensive experiments on multiple databases demonstrate the effectiveness and superiority of our MANet.

\end{abstract}
\begin{IEEEkeywords}
Text-based person search, information inequality, image-specific information, implicit local alignment.
\end{IEEEkeywords}
\IEEEpeerreviewmaketitle
\section{Introduction}
\IEEEPARstart{P}{erson} re-identification (Re-ID) is a hot research topic in computer vision that aims to retrieve a given query from a gallery set collected across cameras. Re-ID can be broadly categorized  into three subtasks based on the query data type: image-based Re-ID~\cite{GLMC, AIESL, cdvr, w1, w7, RAG}, video-based Re-ID~\cite{bicnet, RGSA, yan2018, IFA}, and text-based person search~\cite{GNA}. Over the past decade, significant progress has been made in image-based and video-based Re-ID. Compared to the above two sub-tasks, text-based person search (TBPS) can search the target object with a simpler and more accessible text description. As a result of its practicality, TBPS has gained increasing attention in recent years.
\begin{figure}[t!]
\centering
\subfigbottomskip=-1pt
\subfigcapskip=-1pt
\subfigure[Image background] {\includegraphics[height=0.5in,width=3.2in,angle=0]{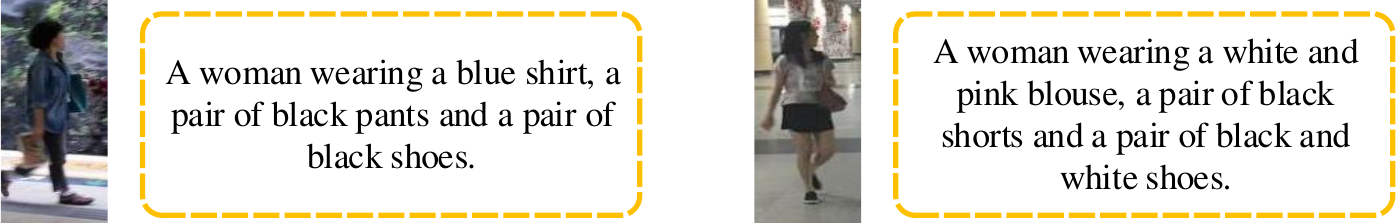}}
\subfigure[Environmental factors] {\includegraphics[height=1.1in,width=3.2in,angle=0]{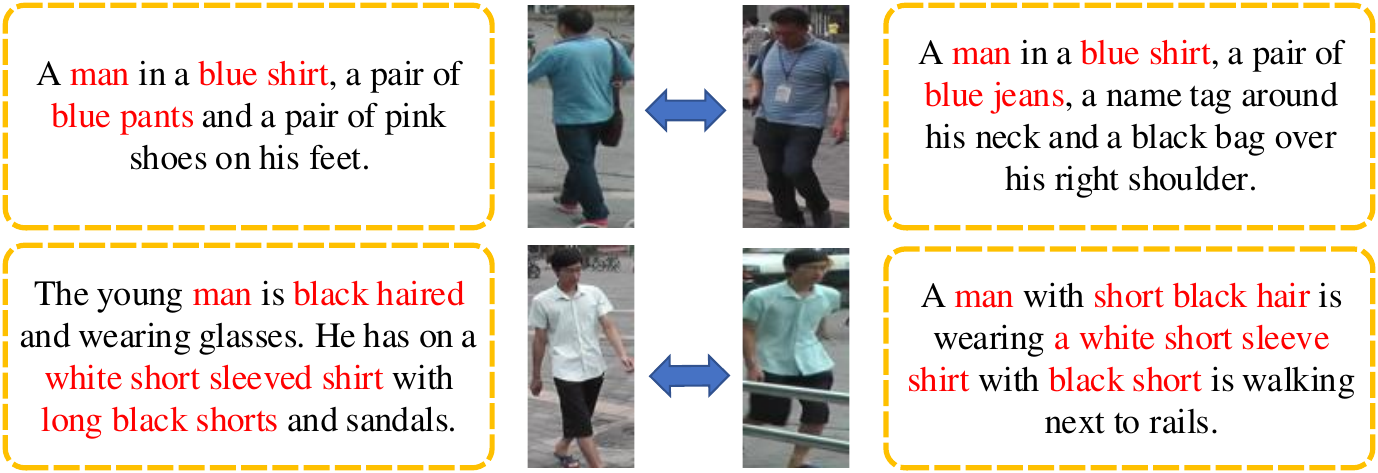}}
\subfigure[Existing explicit local-matching paradigms] {\includegraphics[height=0.8in,width=3.2in,angle=0]{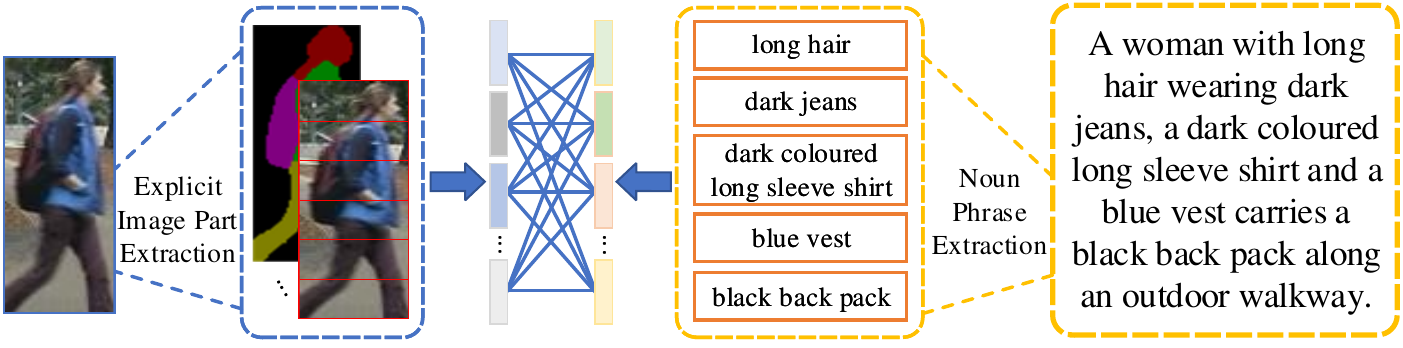}}
\subfigure[Our proposed \textbf{implicit local alignment} paradigm] {\includegraphics[height=0.85in,width=3.2in,angle=0]{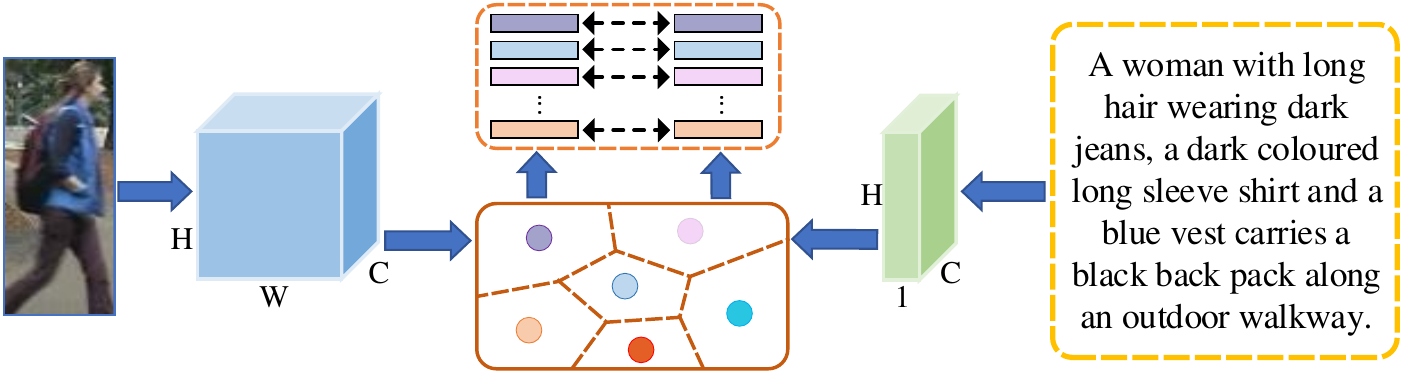}}
\caption{The motivation of our MANet. (a) Cluttered background not described in the text enlarges the gap between modalities. (b) Uncontrollable environmental factors play a negative role to destroy the semantics of the image itself, which are also not described in the text. (c) Existing local-matching methods obtain local parts in an explicit way, and then perform cross-modal interactions to mine local correspondence. (d) Our proposed ILA implicitly learns locally aligned image and text features by a set of learnable modality-shared semantic centers without additional supervision and cross-modal interactions.}
\vspace{-0.35cm}
\label{Fig:1}
\end{figure}

TBPS is challenging as it involves processing information from two heterogeneous modalities, integrating the difficulties of image-based re-ID and image-text matching. The key to TBPS lies in cross-modal alignment. Early methods~\cite{Dual, CMPM, LiTPQT23, CMKA} aligned images and texts into a joint embedding space by designing loss functions or network models. However, these methods focus on aligning global image and text representations while overlooking local fine-grained alignment. Consequently, some later works~\cite{vitaa, CMAAM, NAFS, DSSL, SSAN} have been devoted to modeling local fine-grained correspondence between modalities. The general process involves generating local parts of images/texts, followed by establishing local correspondence through interactions between them, as shown in Figure~\ref{Fig:1}(c). The establishment of fine-grained correspondence depends on the quality of local parts. However, most existing methods directly split images/texts~\cite{NAFS, DSSL, SSAN} or introduce external models~\cite{vitaa, CMAAM} to generate local parts explicitly. As shown in Figure~\ref{Fig:1}, complete and meaningful local parts are irregular, and this hard split way may destroy contextual information, leading to ambiguity when modeling fine-grained correspondence. While the introduction of external models may introduce noise and increase computational costs due to the domain gap between external data and TBPS data. Consequently, such explicitly generated local parts are unreliable for modeling local fine-grained correspondence between modalities. And the direct interactions between modalities may affect each other, which is not conducive to modeling fine-grained correspondence.

Furthermore, most existing methods directly encode images and texts in the same way to obtain image and text features, without considering the information inequality problem caused by modality-specific information in images. Specifically, the text description only includes the appearance information of a pedestrian, while the image captured by the surveillance camera includes additional surrounding environment information not described in the text. This environmental information acts as noise and increases the difficulty of cross-modal alignment, leading to information inequality between images and texts. Our observation suggests that this problem is mainly caused by two types of image-specific information: (1) \textbf{Image background}, as shown in Figure \ref{Fig:1}(a). Cross-modal alignment only focuses on the common information between modalities, while image background that is not described in the text undoubtedly enlarges the gap between modalities. (2) \textbf{Environmental factors}, such as illumination, weather, etc., as shown in Figure \ref{Fig:1}(b), which not only act as noise  but can even destroy the semantics of the image itself, leading to large intra-class variations.

The above observations motivate us to develop a lightweight model that suppresses image-specific information and implicitly learns locally aligned image and text features. To this end, we propose a joint Multi-level Alignment Network (MANet) for TBPS. It comprises an Image-Specific Information Suppression (ISS) module to alleviate the influence of image-specific information and achieve the alignment of information volume between modalities, an Implicit Local Alignment (ILA) and a Global Alignment (GA) module to implicitly learn semantically locally and globally aligned features. 

The ISS module includes a Relation-Guided Localization (RGL) submodule, which calculates the attention weight for each pixel of the image feature map using global relations between the pixel and all other pixels as a guide. With the supervision of the cross-modal alignment loss, our network guided by global relations assigns higher weights to modality-shared human body regions and lower weights to non-pedestrian regions, thereby suppressing background noise. Additionally, we introduce a Channel Attention Filtration (CAF) submodule that performs instance normalization on image features to filter out environmental factor noise followed by recalling identity-related information from the filtered information through channel attention, avoiding the loss of identity information during filtering.

The ILA module implicitly learns locally aligned image and text features by introducing a set of learnable modality-shared semantic topic centers, as shown in Figure \ref{Fig:1}(d). The relations between each pixel (word) feature of the image (text) and these topic centers are used as soft-assign weights to aggregate all pixel (word) features of the image (text) to each semantic center. This generates a set of local image (text) features for a set of centers, which are locally aligned since the semantic centers are modality-shared. The introduction of the modality-shared centers avoids the direct interactions between modalities. Each locally aligned feature is generated based on the complete context information of the image/text, thus avoiding ambiguity and noise, and can better model fine-grained correspondence. The GA module aggregates image salient information spatially and text salient information temporally, maximizing their similarity in joint embedding space for global alignment. The above modules are integrated and jointly optimized in an end-to-end manner. During inference, global and local image/text features extracted by MANet are combined for similarity measurement. 

Our main contributions can be summarized as follows: (1) We design an image-specific information suppression module (including relation-guided localization and channel attention filtration) to address the information inequality between modalities caused by image background and environmental factors. To our best knowledge, we are the first to explicitly clarify the presence of environmental factors for TBPS. (2) An implicit local alignment module is proposed to implicitly learn locally aligned image and text features. (3) we have conducted extensive experiments on two databases to verify the effectiveness of our MANet, it significantly outperforms existing methods.

The remainder of the paper is organized as follows: Section \uppercase\expandafter{\romannumeral2} reviews the works related to the present paper; Section \uppercase\expandafter{\romannumeral3} describes the proposed model in detail, followed by objective function; Section \uppercase\expandafter{\romannumeral4} shows extensive experimental results and analysis; and finally the paper is summarized in Section \uppercase\expandafter{\romannumeral5}.

\section{Related work}
\subsection{Text-based Person Search}
TBPS was first proposed by Li \emph{et al}. \cite{GNA}, and released the first large-scale person description dataset, the CUHK PErson DEScription dataset (CUHK-PEDES), which greatly promoted the development of TBPS. The core challenge in TBPS is achieving cross-modal alignment. To tackle this problem, many methods have been put forward in recent years. Some early works \cite{GNA, IATV, Dual, CMPM, CMKA} primarily focused on global matching between images and texts. For instance, Zhang \emph{et al}.~\cite{CMPM} proposed a cross-modal projection matching loss and a cross-modal projection classification loss to learn modality-shared global features. Chen \emph{et al}.~\cite{CMKA} proposed a cross-modal knowledge adaptation method to solve the information inequality problem between images and texts by directly adapting the knowledge of images to the knowledge of texts at three different levels. Although these global matching-based methods are straightforward and efficient, they may overlook distinctive local details and mix some noise information. 

To mine locally discriminative detail information, a plethora of works \cite{GLA, MIA, PMA, SCAN, vitaa, CMAAM, NAFS, DSSL, SSAN, HGAN, LapsCore} have focused on exploring local matching between images and texts. Some efforts work on more effective attention mechanisms to mine fine-grained correspondence by complex cross-modal pairwise interactions. For example, Niu~\emph{et al}.~\cite{MIA} designed a hierarchically multi-granularity image-text matching mechanism, i.e., global-global, global-local, and local-local, to achieve more accurate similarity evaluation. Afterward, some effective and lightweight models \cite{SSAN, tipcb, SRCF} have been proposed to employ local image parts as a guide to learn correspondingly aligned local text features through cross-modal alignment loss supervision, without complex pairwise interactions. For example, SSAN~\cite{SSAN} designed a word attention module to attend to part-relevant words with the image part feature as a guide. With the success of Transformers in various tasks, several Transformer-based methods~\cite{saf, lgur} have been proposed recently and achieved state-of-the-art performance.

However, most existing local-matching methods rely on explicitly generated local parts to model fine-grained correspondence between modalities. Such explicitly generated local parts may be ambiguous due to the lack of contextual information~\cite{MSCAN, DDAG}, or noisy due to the domain gap between external and TBPS data~\cite{qpm},  which are unreliable for modeling local fine-grained correspondence between modalities. And the direct interactions between modalities may affect each other. In contrast, our method introduces a set of modality-shared semantic centers to implicitly learn locally aligned features, avoiding the guidance or interactions between images and texts. And the generation of each locally aligned feature is based on the complete context information of the image/text. Moreover, several works~\cite{CMKA, SRCF} also consider the information inequality problem between images and texts, but they only paid attention to reducing the impact of image background, ignoring environmental factors. While we suppress image background and filter out environmental factors simultaneously.
\vspace{-0.25cm}

\subsection{Cross-modal Retrieval}
For cross-modal retrieval tasks~\cite{Dual, w2, w5, qda}, the primary challenge is to learn effective and modality-shared visual/textual embeddings in a joint embedding space. Early works~\cite{Dual, CMPM, vse} focus on designing networks and optimization loss to project images and texts into a joint embedding space. For example, Zheng \emph{et al}. \cite{Dual} proposed a dual-path CNN network and an instance loss for image-text retrieval. Zhang \emph{et al}. \cite{CMPM} designed a cross-modal projection matching loss and a cross-modal projection classification loss to learn discriminative image-text embeddings. However, these methods only learn global embeddings from images and texts, ignoring local details and fine-grained correspondence between them. Recently, some advanced works~\cite{SCAN, GSN, t2vlad, SGRAF} have focused on exploring the fine-grained interaction between modalities in the embedding space. For instance, Lee \emph{et al}. \cite{SCAN} employed the bottom-up attention and bi-directional GRU network to get image region features and word features respectively, and presented a stacked cross attention to infer image-text similarity by aligning image region and word features. Liu \emph{et al}. \cite{GSN} proposed a graph-structured matching network to infer fine-grained correspondence between modalities. These methods provide valuable inspiration for TBPS. However, unlike general cross-modal retrieval, all samples for TBPS belong to a single category (person), making it more challenging to mine discriminative details and model fine-grained correspondence. Therefore, this paper aims to explore an effective local alignment framework for TBPS to learn modality-shared local features.
\vspace{-0.25cm}

\subsection{Attention-Based Re-ID}
Re-ID is a challenging task due to various factors such as viewpoint variations, low resolutions, illumination changes, unconstrained poses, occlusions, background clutter, and heterogeneous modalities. The attention mechanism~\cite{w3, w4, w6} have been employed in many re-ID methods \cite{RAG, RGSA, yang2019, IA-Net, mask, nas, MuDeep} to overcome the above challenges due to their ability to focus on important information related to the task and suppress irrelevant information. For instance, Yang \emph{et al}. \cite{yang2019} designed intra-attention and inter-attention modules to learn robust and discriminative human features from global image and local parts, respectively. To introduce global structural information, Zhang \emph{et al}. \cite{RAG} and Li \emph{et al}. \cite{RGSA} explored global scope relations for feature nodes at each location to learn attention for informative and discriminative feature learning. To eliminate the influence of background clutter on re-ID, Song \emph{et al}. \cite{mask} designed a mask-guided contrastive attention module by introducing external clues to guide the network to focus on the human body region. Inspired by these works, we design two attention modules to address the information inequality problem between images and texts. The attention modules guide the network to focus on the modality-shared human body region in images and filter out environmental factors while preserving identity information.

\begin{figure*}[t!]
  \centering
  \setlength{\abovecaptionskip}{-2pt}
  \includegraphics[width=6.8in,height=4.1in]{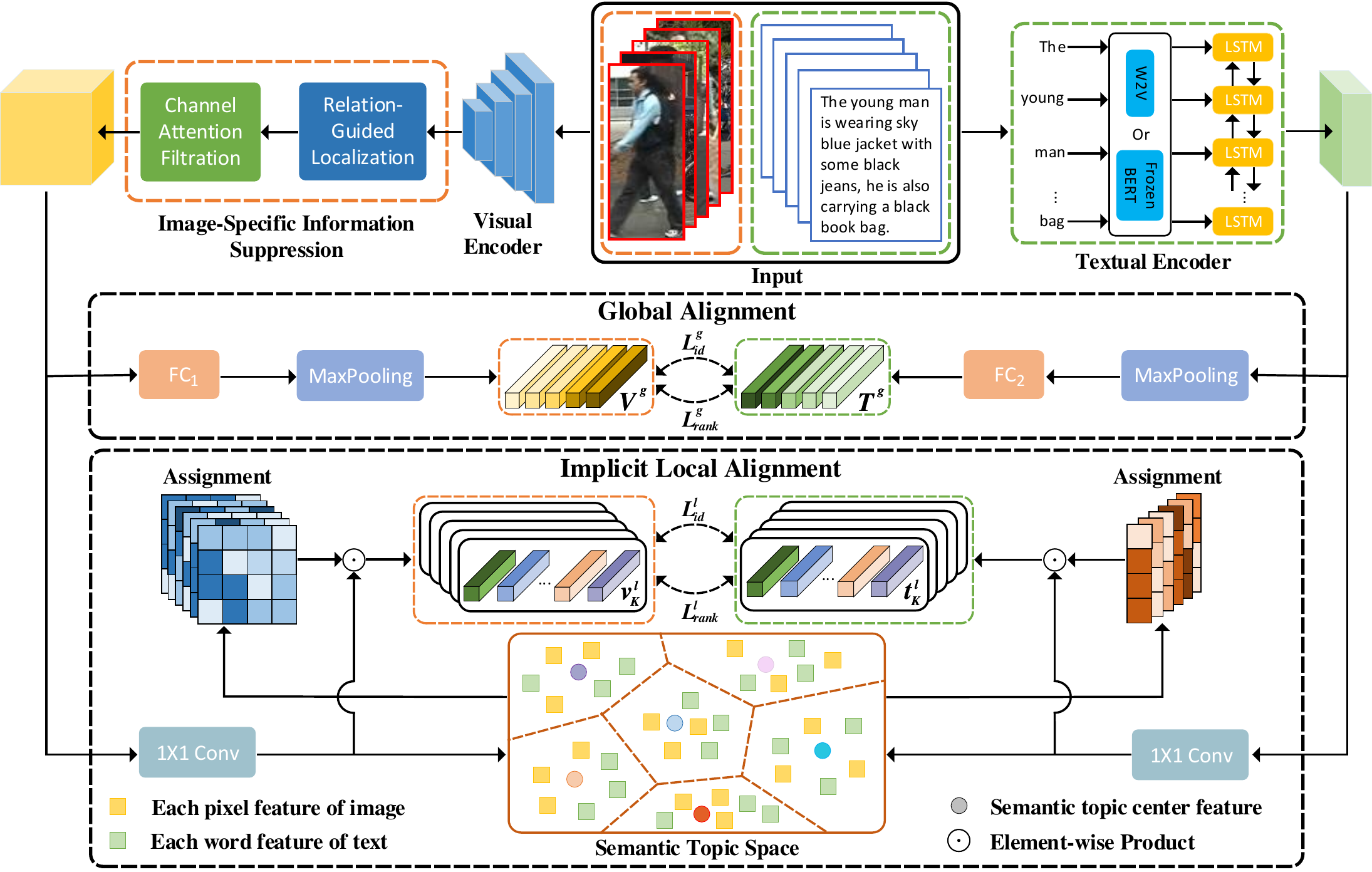}\\
  \caption{Overview of the proposed MANet. Given image-text pairs, we first extract image and text feature maps by image and text encoder respectively. Next, we feed the image feature map to an image noise suppression module to eliminate the effects of the image background and environmental factors through relation-guided localization and channel attention filtration respectively. Finally, the output image and text features are sent to the global and local alignment branches respectively. In the global branch, the image and text features are respectively encoded into a joint embedding space using two fully connected layers for global alignment. In the local branch, we design a set of modality-shared learnable semantic topic centers and calculate the assignment of the pixel (word) features of the image (text) to each center. we aggregate the image and text features based on the assignments to generate the locally aligned image and text features. The generated global and local features for images and texts are supervised by Ranking loss and Identification loss to achieve multi-level modality alignment.}
  \vspace{-0.35cm}
  \label{Fig:2}
\end{figure*}

\section{MANet Framework}
In this section, we elaborate on the implementation details of our MANet framework, and the overview is shown in Figure \ref{Fig:2}. We begin by extracting visual and textual features using CNN and RNN networks, respectively. Next, we feed the extracted image features to the image-specific information suppression module, which includes relation-guided localization and channel attention filtration to eliminate the influence of image background and environmental factors. The processed image and text features are then fed to the global and local encoding branches for global and local alignment, respectively. Finally, the above modules are jointly optimized by identification loss, cross-modality ranking loss, and identity-relevant content consistency loss to learn modality-aligned features for text-to-image retrieval.
\vspace{-0.25cm}
\subsection{Visual and Textual Representations}
For a batch of $N$ image-text pairs $\{I_i, T_i\}_{i=1}^N$ and corresponding ground-truth label set $\{\bm y_{i}\}_{i=1}^N$ drawn from a TBPS database, where each $\bm y_{i}\in \mathbb{R}^{N_y}$ is a one-hot ground-truth label vector, $N_y$ is the number of identities. Given a text-image pair $(I_{i}, T_{i})$, the goal is to encode it into a joint feature space to measure the similarity. For an image $I_{i}$, we utilize a ResNet-50 \cite{resnet} pre-trained on ImageNet \cite{imagenet} with the final fully-connected layer removed as the visual backbone to extract the visual feature map $\bm F\in \mathbb{R}^{C\times H\times W}$, where $C$ denotes the number of feature channel with height $H$, and width $W$. Our goal is to achieve both global and local alignment between images and texts. To obtain the global and local features for each image, we introduce the global image branch $F^{image}_{glb}$ and the local branch $F_{loc}$ to generate a global visual feature $\bm v^{g}=F^{image}_{glb}(\bm F)\in \mathbb{R}^{d_g}$ and a set of local visual features $\{\bm v^{l}_{1}, \bm v^{l}_{2}, ..., \bm v^{l}_{K}\}=F_{loc}(\bm F)\in \mathbb{R}^{K\times d_c}$, where $K$ denote the number of local features.

Given a textual caption $T_{i}$, we first encode each word as a vector, i.e. word embedding. Two commonly used methods for TBPS are to generate word embedding through Word2Vector (W2V)~\cite{SSAN} or BERT~\cite{bert}. For a fair comparison, we generate word embedding in both ways. For W2V, we first create a vocabulary from the text database of the training set, only counting some popular words whose appearance frequency is more than 2 \cite{SSAN}, and the size of the vocabulary is $V$. After that, we first encode each word in $T_{i}$ into a length-$V$ one-hot vector $\bm v_j$. Then we transform the one-hot vector $\bm v_j$ by a word embedding matrix $\bm W_e\in \mathbb{R}^{V\times d_e}$ into a $d_e$-dimensional word embedding $\bm w_{j}=\bm W_{e}\bm v_j$. For BERT, we first use the lower-cased byte pair encoding (BPE) with a 30522 vocabulary size to tokenize the text $T_{i}$. Then, the textual token sequence is fed into the pre-trained BERT model to generate the word embedding. Note that BERT is only used to generate word embedding, which is frozen during training.

The long short-term memory network (LSTM) as a special RNN has shown great capabilities in capturing the long-term dependencies among words. We adopt bi-directional LSTM (bi-LSTM) as a text backbone to extract text features. After the embedding stage, all word embeddings $\{\bm w_{1}, \bm w_{2}, ..., \bm w_{L}\}\in \mathbb{R}^{d_e\times L}$ of the caption $T_{i}$ generated by W2V or BERT is feed into the bi-LSTM network in sequence to get the hidden states of forward and backward directions for each word. By averaging the hidden states of two directions to generate the corresponding contextual-aware text representation $\bm e_i\in \mathbb{R}^{C}$ for each word. All word representations for a caption are stacked as $\bm E=[\bm e_{1}, \bm e_{2}, ..., \bm e_{L}]\in \mathbb{R}^{C\times L}$, where $L$ is the number of words in each caption. It should be noted that the length of each caption is different. To ensure the consistency of caption length, we unify the length of the caption as $L$, select the first $L$ words when the length is greater than $L$, and fill zero at the end of the caption when the length is less than $L$.

Similarly, in order to perform global and local alignment, we also obtain the global text feature $\bm t^{g}=F^{text}_{glb}(\bm E)\in \mathbb{R}^{d_g}$ and local text features $\{\bm t^{l}_{1}, \bm t^{l}_{2}, ..., \bm t^{l}_{K}\}=F_{loc}(\bm E)\in \mathbb{R}^{K\times d_c}$ for each caption by the global text branch $F^{text}_{glb}$ and the local branch $F_{loc}$. The details of the global branches $F^{image}_{glb}$, $F^{text}_{glb}$ and the local branch $F_{loc}$ are presented in the following.
\vspace{-0.25cm}

\subsection{Image-Specific Information Suppression}
Considering the fact that the information contained in an image and a text is not equal, it is embodied in two aspects: Firstly, the image is captured at a distance from the person and encompasses not only the pedestrian but also the surrounding background (Figure \ref{Fig:1}(a)). Secondly, the Re-ID task involves capturing images from multiple cameras at different times and places. Due to the differences in camera parameters and environment (lighting conditions, weather, viewpoint, etc.), the captured images encompass environmental factors that lead to significant intra-class variations, meaning multiple images of the same person are very different (Figure \ref{Fig:1}(b)). Texts are generally cleaner than images, as they mostly describe pedestrian appearance information. Therefore, the background and environmental factors~\cite{TangYLT22, ZhaTST23} in the image act as noise for text, which increases the modality gap between images and texts, leading to the information inequality problem. To address this issue, we propose an image-specific information suppression module consisting of two submodules: relation-guided localization (RGL) and channel attention filtration (CAF), to deal with background and environmental factors respectively. The solution to this problem can effectively narrow the modality gap and achieve the alignment of information volume between modalities.

\begin{figure}[t!]
  \centering
  \setlength{\abovecaptionskip}{-1pt}
  \includegraphics[width=3.5in,height=1.7in]{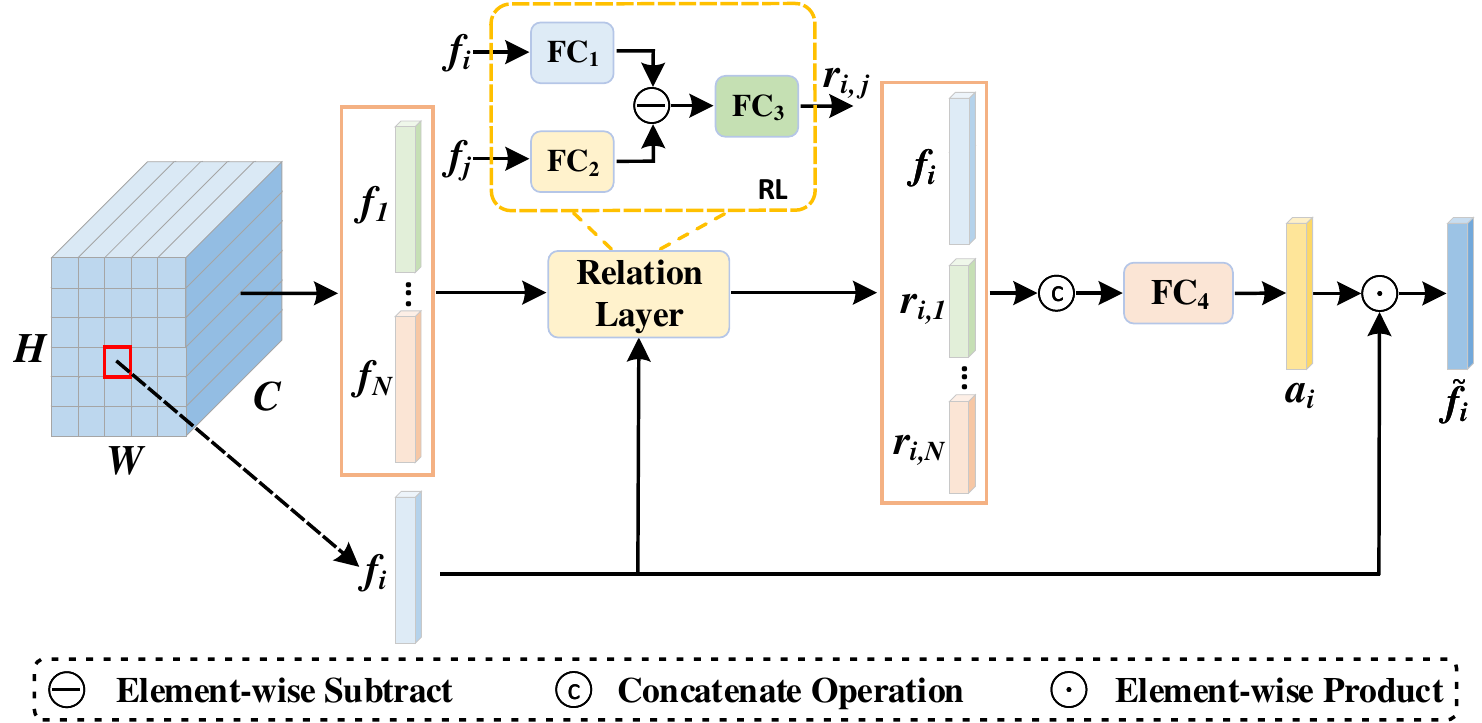}\\
  \caption{The architecture of relation-guided localization (RGL) module.} \label{Fig:3}
  \vspace{-0.5cm}
\end{figure}

\textbf{Relation-Guided Localization (RGL)}. The attention mechanism is widely used in various tasks due to its ability to enhance discriminative features and suppress irrelevant ones, which is in line with our goal. Zhang~\emph{et al}.~\cite{RAG} has proved that the relations among spatial positions in the image provide clustering-like information, which is helpful for the semantic understanding of the image. Therefore, a relation-guided localization module is introduced to guide attention learning through the relations among spatial positions. This enables the module to suppress the background and highlight the pedestrian region in the image that corresponds to the text description, as illustrated in Figure \ref{Fig:3}.

It is crucial how to measure the relation between features effectively. The most common way is to use the Euclidean or cosine distance to calculate a relational scalar between different features, which indicates the degree of correlation between them while lacking detailed correspondence~\cite{TangLPT20}. Unlike them, we capture more detailed associations among spatial positions by computing a relation vector instead of a scalar. The relation function between vectors $\bm x\in \mathbb{R}^{C}$ and $\bm y\in \mathbb{R}^{C}$ is formulate as:
\begin{equation}\
\begin{aligned}
r(\bm x, \bm y; \bm W, \bm W_\theta, \bm W_\phi)=ReLU(BN(\bm W (\bm x_\theta-\bm y_\phi))
\end{aligned},
\end{equation}
where $\bm W\in \mathbb{R}^{\frac{C}{r_2}\times \frac{C}{r_1}}$ denotes a weight matrix of a fully connected (FC) layer, $BN(\cdot)$ is a batch normalization (BN) layer and $ReLU(\cdot)$ is a rectified linear unit activation layer. Based on this, we can get a $\frac{C}{r_2}$-dimensional relation vector. $\bm x_\theta=ReLU(BN(\bm W_\theta \bm x))$ and $\bm y_\phi=ReLU(BN(\bm W_\phi \bm y))$. $\bm W_\theta$, $\bm W_\phi\in \mathbb{R}^{\frac{C}{r_1}\times C}$ reduce the feature dimension with factor $r_1$.

With the visual feature map $\bm F\in \mathbb{R}^{C\times H\times W}$ from the visual backbone, we calculate the relations between each spatial location and all spatial locations to obtain the attention score of each spatial location. Specifically, for a feature vector $\bm f_i\in \mathbb{R}^{C}$ at $i$-th position, we compute the relation between it and all $N (N=H\times W)$ positions to get $N$ relation vectors. Then the $N$ relation vectors are concatenated to denote the global relation vector $\bm r_i\in \mathbb{R}^{\frac{NC}{r_2}}$at $i$-th position:
\begin{equation}\
\begin{aligned}
\bm r_i=Concat([\bm r_{i,1},\bm r_{i,1}, ..., \bm r_{i,N}])\\
\end{aligned},
\end{equation}
\begin{equation}\
\begin{aligned}
\bm r_{i,j}=r(\bm f_i, \bm f_j, \bm W, \bm W_\theta, \bm W_\phi), (1 \leq j \leq N)
\end{aligned}.
\end{equation}

To obtain the attention at the $i$-th position, we combine the visual feature $\bm f_i$ with its corresponding global relation vector $\bm r_i$ to generate the attention by a projection function implemented by FC layer followed by BN and Sigmoid function:
\begin{equation}\
\begin{aligned}
\bm a_i=Sigmoid(BN(\bm W_a [\bm f_i, \bm r_i]))
\end{aligned},
\end{equation}
where $\bm W_a\in \mathbb{R}^{C\times(1+\frac{N}{r_2})C}$ and $\bm a_i\in \mathbb{R}^{C}$. For all positions, we get an attention map $\bm A\in \mathbb{R}^{C\times H\times W}$. Finally, the attention map $\bm A$ and the visual feature map $\bm F$ perform element-wise product to get the enhanced feature map $\widetilde{\bm F}$. The enhanced feature map $\widetilde{\bm F}$ is subsequently processed in the later stage and fed into the joint space to align with the corresponding text. As the text does not provide any information about the image background, under the supervision of cross-modal alignment loss, the network guided by global relations assigns higher weights to modality-shared human body regions and lower weights to non-pedestrian regions to suppress the background noise.

\begin{figure}[!t]
  \centering
  \setlength{\abovecaptionskip}{-1pt}
  \includegraphics[width=3.4in,height=1.3in]{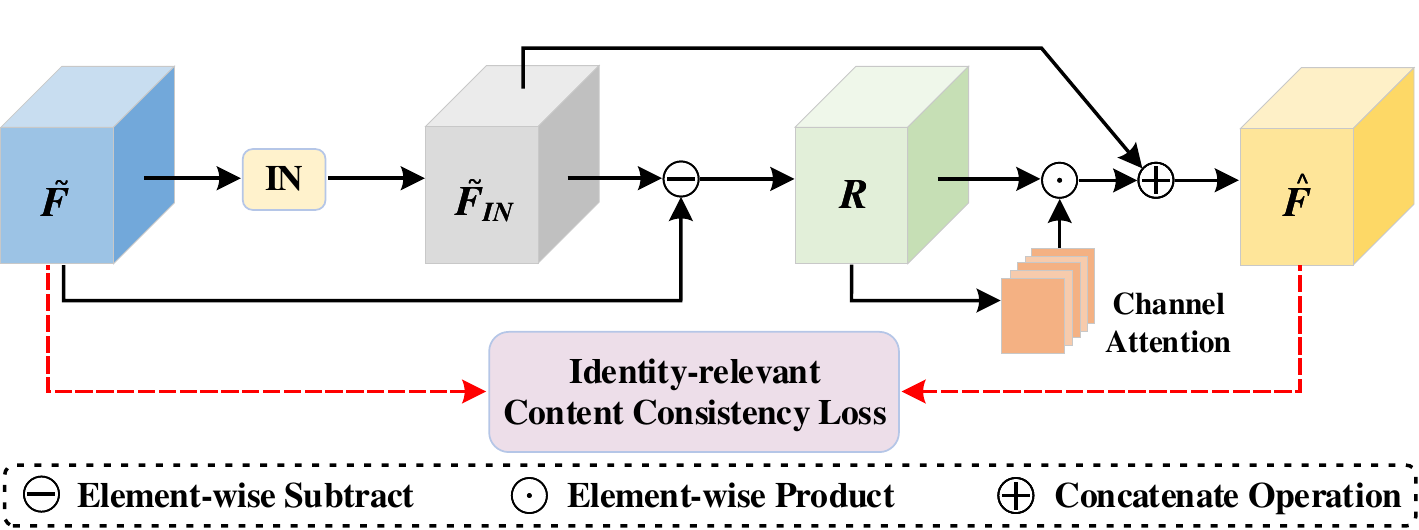}\\
  \caption{The architecture of channel attention filtration (CAF) module.}
  \vspace{-0.55cm}
  \label{Fig:4}
\end{figure}

\textbf{Channel Attention Filtration (CAF)}. The presence of environmental factors in images not only widens the gap between modalities but also results in significant intra-class variations within images. To eliminate the influence of environmental factors, we introduce a channel attention filtration module, depicted in Figure \ref{Fig:4}.

Previous studies~\cite{arb,snr, two} have shown that Instance Normalization (IN) can effectively reduce the discrepancy among instances by filtering out instance-specific style information. However, directly applying IN can potentially remove some identity-relevant information, thereby damaging the discrimination capability of features. To avoid this issue, we further mine the identity-relevant information from the IN-removed information after instance normalization. Specifically, we apply Instance Normalization to the visual feature $\widetilde{\bm F}$ processed by RGL to remove the style information.
\begin{equation}\
\begin{aligned}
\widetilde{\bm F}_{IN}=IN(\widetilde{\bm F})=\gamma(\frac{\widetilde{\bm F}-\mu(\widetilde{\bm F})}{\sigma(\widetilde{\bm F})})+\beta
\end{aligned},
\end{equation}
where $\mu(\cdot)$ and $\sigma(\cdot)$ denote the mean and standard-deviation, $\gamma, \beta\in \mathbb{R}^{C}$ are parameters learned from data. Next, we restitute the identity-relevant information by distilling it from the IN-removed information. The IN-removed information is defined as:
\begin{equation}\
\begin{aligned}
\bm R=\widetilde{\bm F}-\widetilde{\bm F}_{IN}
\end{aligned}.
\end{equation}

With the IN-removed information $\bm R$, we mine the identity-relevant information from it through channel attention and then add it to the instance normalized feature.
\begin{equation}\
\begin{aligned}
\hat{\bm F}=\widetilde{\bm F}_{IN}+\bm w_C\odot\bm R
\end{aligned},
\end{equation}
where $\odot$ denotes the element-wise product operation, $\bm w_C\in \mathbb{R}^{C}$ is the channel-wise attention weight indicating the identity-relevant channels, which is generated in the same way as \cite{senet}. With the channel attention filtration module, we expect that the output feature can fully retain effective identity information while discarding redundant noise information. To ensure this, we introduce an identity-relevant content consistency loss, which is explained in detail in~\ref{sec.E}.

After text encoding, image encoding, localization, and filtration, we alleviate the information inequality problem between images and texts, resulting in obtaining the text feature map $\bm E\in \mathbb{R}^{C\times L}$ and image feature map $\hat{\bm F}\in \mathbb{R}^{C\times H\times W}$. Although $\hat{\bm F}$ and $\bm E$ achieve the alignment of information volume, images are spatially structured while texts are orderless. Directly comparing these two heterogeneous features is not feasible as they are not well-aligned at the semantic level. Therefore, we perform global and local semantic alignment between images and texts in the following sections.
\vspace{-0.6cm}

\subsection{Implicit Local Alignment}
Local-level alignment plays a crucial role in TBPS as it allows for the mining of fine-grained correspondences between images and texts. Pedestrian expressions can take various forms, such as image, text, and voice, but their underlying semantic information remains the same. Therefore, we hypothesize that there exists a set of latent semantic topic centers in a shared semantic space, which can completely express the semantic information of pedestrians and are shared across modalities. These centers can be compared to atoms in a dictionary, where different combinations of atoms can express anything. Building on this idea, we propose an implicit local alignment (ILA) module that adaptively aggregates all pixel/word features of image/text to the same centers, resulting in multiple locally aligned image/text features.

Assuming the existence of a set of semantic topic centers $\{\bm c_1, \bm c_2, ..., \bm c_K\}\in \mathbb{R}^{K\times d_c}$ in a shared semantic space, where $K$ denotes the number of centers and $\bm c_i\in \mathbb{R}^{d_c}$, $d_c$ is the dimension of shared semantic space. These centers can be learned jointly with the entire network. To assign pixel/word features of image/text to their respective topic centers, we calculate the relations between the pixel (word) features of the image (text) and the centers, similar to Eq. (1). We provide a detailed explanation of the assignment on visual feature map, followed by a brief description of the text feature map, as the operation is the same for both feature maps. Given the visual feature map $\hat{\bm F}\in \mathbb{R}^{C\times H\times W}$, first project it into the shared semantic space by a $1\times 1$ convolution layer $\bm W_s\in \mathbb{R}^{d_c\times C}$ after L2-normalization:
\begin{equation}\
\begin{aligned}
\bm Z^{image}=\bm W_s Normalize(\hat{\bm F})
\end{aligned},
\end{equation}
where $\bm Z^{image}\in \mathbb{R}^{d_c\times H\times W}$. Next, we compute assignments on the corresponding centers for each pixel feature. Specifically, for the feature vector $\bm z_i^{image}\in \mathbb{R}^{d_c}$ at $i$-th pixel in $\bm Z^{image}$, its assignment to $j$-th topic center $\bm c_j$ can be generated in the same way as Eq. (1) as follow:
\begin{equation}\
\begin{aligned}
\bm a_{i,j}^{image}=r(\bm z_i^{image}, \bm c_j, \bm W^l, \bm W_{\theta}^l, \bm W_{\phi}^l)
\end{aligned},
\end{equation}
where $\bm W^l\in \mathbb{R}^{d_c\times \frac{d_c}{r_3}}$, and $\bm W_{\theta}^l$, $\bm W_{\phi}^l\in \mathbb{R}^{\frac{d_c}{r_3}\times d_c}$ are all learnable parameter matrices. Then the aggregated local image feature for each center can be obtained,
\begin{equation}\
\begin{aligned}
\bm v_j^{l}=\sum_{i=1}^{N_s}\bm a_{i,j}^{image}\odot\bm z_i^{image}
\end{aligned},
\end{equation}
where $\odot$ denotes the element-wise product operation, $N_s=H\times W$ denotes the number of pixels. Thus, we can generate a set of aggregated local image features $\{\bm v_1^{l}, \bm v_2^{l}, ..., \bm v_K^{l}\}$ for all semantic topic centers. Similarly, for the textual feature map $\bm E$, we first extend $\bm E\in \mathbb{R}^{C\times L}$ to $\hat{\bm E}\in \mathbb{R}^{C\times L\times 1}$ to meet the input requirement of convolutional layer, and then generate a set of aggregated local text features $\{\bm t_1^{l}, \bm t_2^{l}, ..., \bm t_K^{l}\}$ in the same way using the share semantic topic centers.
\begin{equation}\
\begin{aligned}
\bm Z^{text}=\bm W_s Normalize(\hat{\bm E})
\end{aligned},
\end{equation}
\begin{equation}\
\begin{aligned}
\bm a_{i,j}^{text}=r(\bm z_i^{text}, \bm c_j, \bm W^l, \bm W_{\theta}^l, \bm W_{\phi}^l)
\end{aligned},
\end{equation}
\begin{equation}\
\begin{aligned}
\bm t_j^{l}=\sum_{i=1}^{L}\bm a_{i,j}^{text}\odot\bm z_i^{text}
\end{aligned},
\end{equation}
where $L$ denotes the length of caption, $\bm z_i^{text}\in \mathbb{R}^{d_c}$ is the feature vector of $i$-th word in $\bm Z^{text}$. Finally, we respectively concatenate all local image and text features to get the output feature $\bm v^l\in \mathbb{R}^{Kd_c}$ and $\bm t^l\in \mathbb{R}^{Kd_c}$, which are aligned effectively since the learned local features for images and texts share the same semantic topic centers.
\vspace{-0.5cm}

\subsection{Global Alignment}
Moreover, we also introduce global alignment since the global features contain more comprehensive information and are complementary to local features. For visual feature map $\hat{\bm F}\in \mathbb{R}^{C\times H\times W}$, we first project it into a joint semantic embedding space through a fully connected layer $\bm W_g^{image}$. Because only part of the image region is related to the text description, considering that the global max-pooling layer can mine salient information and filter out noise, thus perform global max-pooling (GMP) to obtain the global visual feature.
\begin{equation}\
\begin{aligned}
\bm v^{g}=GMP(\bm W_g^{image}\hat{\bm F})
\end{aligned}.
\end{equation}

For text feature map $\hat{\bm E}\in \mathbb{R}^{C\times L\times 1}$, we first perform global max-pooling to get the feature vector, followed by projecting the feature vector into the joint semantic embedding space by a fully connected layer $\bm W_g^{text}$ to obtain the global text feature.
\begin{equation}\
\begin{aligned}
\bm t^{g}=\bm W_g^{text}GMP(\hat{\bm E})
\end{aligned},
\end{equation}
where $\bm v^g$, $\bm t^g\in \mathbb{R}^{d_g}$. For a batch of $N$ image-text pairs $\{I_i, T_i\}_{i=1}^N$, we generate the global image and text feature set, $\bm V^g$, $\bm T^g\in \mathbb{R}^{N\times d_g}$, respectively.
\vspace{-0.25cm}

\subsection{Training Loss and Inference}\label{sec.E}
After obtaining the global and local features for images and texts using the proposed model, we train the model with identification loss, cross-modality triplet ranking loss, and identity-relevant content consistency loss. The detailed descriptions of each loss are presented in the following.

\textbf{Identification Loss}. We adopt the identification loss to classify persons into different groups by their identities, which ensures identity-level matching. Specifically, for image-text pair $(I_i, T_i)$, the cross-entropy loss is imposed on the global feature ($\bm v^g$, $\bm t^g$) and each of the $K$ local features ($\{\bm v_k^{l}\}_{k=1}^K$, $\{\bm t_k^{l}\}_{k=1}^K$). Therefore, the total identification loss $\mathcal{L}_{id}$ is:
\begin{equation}\
\begin{aligned}
\mathcal{L}_{id}=\mathcal{L}_{id}^{g}+\sum_{k=1}^{K}\mathcal{L}_{id}^{l_k}
\end{aligned},
\end{equation}

\begin{equation}\
\begin{aligned}
\mathcal{L}_{id}^{g}=&-y_i log(softmax((\bm W^{id}_g)^T\bm v^g))\\
                     &-y_i log(softmax((\bm W^{id}_g)^T\bm t^g))
\end{aligned},
\end{equation}

\begin{equation}\
\begin{aligned}
\mathcal{L}_{id}^{l_k}=&-y_i log(softmax((\bm W^{id}_{l_k})^T\bm v_k^{l}))\\
                       &-y_i log(softmax((\bm W^{id}_{l_k})^T\bm t_k^{l}))
\end{aligned},
\end{equation}
where $\bm W^{id}_g\in \mathbb{R}^{d_g\times N_y}$, $\bm W^{id}_{l_k}\in \mathbb{R}^{d_c\times N_y}$ are the parameters of the fully connected layer for classification, $y_i$ is the one-hot ground-truth label vector of image $I_i$ (text $T_i$).

\textbf{Cross-Modality Ranking Loss}. We employ the common cross-modality bi-directional dual-constrained ranking loss to supervise the modality-shared global and local feature learning. In addition, we also introduce text descriptions of other images with the same identity as \cite{SSAN} as weak supervision signals to overcome the intra-class variance in the text.
\begin{equation}\
\begin{aligned}
~~~~~~~~\mathcal{L}_{rank}^{z}&=max(\alpha_1-S(\bm v^z_p, \bm t^z_p)+S(\bm v^z_p, \bm t^z_n), 0)\\
&+max(\alpha_1-S(\bm v^z_p, \bm t^z_p)+S(\bm v^z_n, \bm t^z_p), 0)\\
&+\lambda_1 max(\alpha_2-S(\bm v^z_p, \overline{\bm t}^z_p)+S(\bm v^z_p, \bm t^z_n), 0)\\
&+\lambda_1 max(\alpha_2-S(\bm v^z_p, \overline{\bm t}^z_p)+S(\bm v^z_n, \overline{\bm t}^z_p), 0)
\end{aligned},
\end{equation}
where $\bm v^z$ and $\bm t^z (z=g, l)$ denote the global or local feature for image and text, respectively. $\overline{\bm t}^z$ denotes the text feature corresponding to another image with the same identity as $\bm v^z$. $(\bm v^z_p, \bm t^z_p)$ denotes the matched image-text pairs, and $(\bm v^z_p, \bm t^z_n), (\bm v^z_n, \bm t^z_p)$ denote the mismatched pairs. $\alpha_1$ and $\alpha_2$ indicate the margins, and the setting strategy of $\alpha_2$ keeps same with \cite{SSAN}, $\lambda_1$ is a hype-parameter.
\begin{equation}\
\begin{aligned}
\mathcal{L}_{rank}=\mathcal{L}_{rank}^{g}+\mathcal{L}_{rank}^{l}
\end{aligned}.
\end{equation}

\textbf{Identity-relevant Content Consistency Loss} is introduced to ensure that CAF can fully retain useful identity information while discarding redundant noise information. Specifically, we hope that the identity-relevant information can remain consistent before and after passing through CAF, while only filtering out noise information. A triplet loss is adapted to achieve the goal.
\begin{equation}\
\begin{aligned}
\mathcal{L}_{cons}&=max(\alpha_3-S(\widetilde{\bm f}, \hat{\bm f})+S(\widetilde{\bm f}, \hat{\bm f}_n), 0)\\
&+max(\alpha_3-S(\widetilde{\bm f}, \hat{\bm f})+S(\widetilde{\bm f}_n, \hat{\bm f}), 0)
\end{aligned},
\end{equation}
where $\widetilde{\bm f}=GAP(\widetilde{\bm F})$, $\hat{\bm f}=GAP(\hat{\bm F})$ respectively represent the feature vectors before and after $\bm F$ pass through CAF, and $GAP(\cdot)$ denotes the global average pooling. $\widetilde{\bm f}_n$, $\hat{\bm f}_n$ denote the hardest negative sample for $\bm F$ before and after through CAF in a mini-batch, respectively. $\alpha_3$ is the corresponding margin. $S(\cdot, \cdot)$ denotes the similarity function. The loss retains the identity information by pulling the distance between $\widetilde{\bm f}$ and $\hat{\bm f}$ as close as possible and ensures the discrimination ability of features by pulling distances between $\widetilde{\bm f}$ and $\hat{\bm f}_n$, $\widetilde{\bm f}_n$ and $\hat{\bm f}$ as far as possible.

Finally, the total loss $\mathcal{L}_{total}$ is utilized to train our model in an end-to-end manner as follow:
\begin{equation}\
\begin{aligned}
\mathcal{L}_{total}=\mathcal{L}_{id}+\mathcal{L}_{rank}+\lambda_2\mathcal{L}_{cons}
\end{aligned},
\end{equation}
where $\lambda_2$ is a hype-parameter to balance the contributions of individual loss terms.

During inference, we compute the similarity scores $S^g$, $S^l$ of global and local features between each text query and image gallery, respectively, using the cosine similarity function. Based on this score $S=S^g+S^l$, we rank the person images in the image gallery and retrieve the top matches corresponding to the text query.
\vspace{-0.2cm}

\section{Experiments}
In this section, we conduct extensive experiments on two TBPS databases to verify the effectiveness and superiority of our MANet. We first introduce two large-scale public databases, the implementation and training details of our proposed model. Next, we compare our method with state-of-the-art methods on both databases. Finally, we present some ablation studies to demonstrate the effectiveness of each component of our MANet.
\vspace{-0.65cm}

\begin{table}[!ht]\small
\setlength{\abovecaptionskip}{0.1cm}
\centering {\caption{Performance comparison with state-of-the-art methods on CUHK-PEDES. R@1, R@5, and R@10 are listed. 'WE' means word embedding via W2V or BERT. '-' denotes that no reported result is available.}\label{Tab:1}
\renewcommand\arraystretch{1.2}
\begin{tabular}{m{2.0cm}<{\centering}|m{0.4cm}<{\centering}|m{1.3cm}<{\centering}|m{0.7cm}<{\centering}m{0.7cm}<{\centering}m{0.7cm}<{\centering}}
\hline
 \hline
  Methods & WE &Ref & R@1 & R@5 & R@10\\
  \hline
  A-GANet~\cite{A-GANet} & \multirow{15}{*}{\rotatebox{90}{W2V}}  & MM19 & 53.14 & 74.03 & 81.95  \\ 
  
  Dual Path~\cite{Dual} &  & TOMM20 & 44.40 & 66.26 & 75.07  \\

  MIA~\cite{MIA} &  & TIP20 & 53.10 & 75.00 & 82.90  \\

  PMA~\cite{PMA}  & & AAAI20 & 53.81 & 73.54 & 81.23  \\

  TDE~\cite{TDE} & & MM20 & 55.25 & 77.46 & 84.56  \\

  ViTAA~\cite{vitaa} &  & ECCV20 & 55.97 & 75.84 & 83.52  \\

  IMG-Net~\cite{IMG-Net} & & JEI20 & 56.48 & 76.89 & 85.01  \\

  CMAAM~\cite{CMAAM} & & WACV20 & 56.68 & 77.18 & 84.86  \\

  CMKA~\cite{CMKA} &  & TIP21 & 54.69  &73.65  &81.86 \\

  DSSL~\cite{DSSL}  & & MM21 & 59.98  &80.41  &87.56 \\
  
  MGEL~\cite{mgel} & & IJCAI21 & 60.27  &80.01  &86.74 \\

  SSAN~\cite{SSAN} &  & arXiv21 & 61.37  &80.15  &86.73 \\

  SUM~\cite{sum} &  & KBS22 & 59.22  & 80.35  & 87.60  \\

  \hline

  TIMAM~\cite{TIMAM} & \multirow{15}{*}{\rotatebox{90}{BERT}} & ICCV19 & 54.51 & 77.56 & 84.78  \\

  HGAN~\cite{HGAN} &   & MM20 & 59.00 & 79.49 & 86.62  \\

  NAFS~\cite{NAFS}  &  & arXiv21 & 59.94  &79.86  &86.70 \\

  LapsCore~\cite{LapsCore} &  & ICCV21 & 63.40 & - & 87.80  \\

  ACSA~\cite{ACSA} &   & TMM22  & 63.56   & 81.40  & 87.70 \\
  
  IVT~\cite{ivt} &   & ECCVW22 & 64.00  & 82.72   & 88.95 \\

  SRCF~\cite{SRCF} &   & ECCV22 & 64.04   & 82.99   & 88.81    \\

  LBUL~\cite{LBUL} &   & MM22 & 64.04  & 82.66  & 87.22    \\
  
  SAF~\cite{saf}  &   & ICASSP22 & 64.13  & 82.62  & 88.40 \\

  TIPCB~\cite{tipcb}  &   & Neuro22 & 64.26  & \underline{83.19}  & \underline{89.10} \\
  
  CAIBC~\cite{caibc} &   & MM22 & 64.43  & 82.87   & 88.37 \\
  
  AXM-Net~\cite{AXM-Net} &   & AAAI22 & 64.44  & 80.52  & 86.77 \\ 

  C$_2$A$_2$~\cite{C2A2} &   & MM22 & 64.82  & \bf{83.54}  & \textbf{89.77} \\ 

  LGUR~\cite{lgur}  &   & MM22  & \underline{65.25}  & 83.12  & 89.00    \\

  RKT~\cite{RKT}  &   & TMM23 & 61.48  &80.74  &87.28 \\

  \hline
  \textbf{MANet-W2V}  & & - & 63.92 & 82.15 & 87.69  \\
  \textbf{MANet-BERT}  & & - & \bf{65.64} & 83.01 & 88.78  \\
  \hline\hline
\end{tabular}}
\vspace{-0.5cm}
\end{table}

\subsection{Experiment Settings}
\subsubsection{Datasets}
\textbf{CUHK-PEDES}~\cite{GNA} comprises 40,206 images and 80,412 text descriptions of 13,003 persons. Each image is manually annotated with 2 text descriptions with an average length of not less than 23 words. The database constitutes a vocabulary with 9408 unique words. To enable a fair comparison with existing methods, we follow the official data split protocol in \cite{GNA}. Specifically, 34,054 images of 11,003 persons and corresponding 68,108 text descriptions are used as the training set. The remaining 2000 persons are equally divided into the validation and testing sets, with the validation set comprising 3,078 images and 6,156 text descriptions, and the testing set comprising 3,074 images and 6,148 text descriptions.

\textbf{ICFG-PEDES}~\cite{SSAN} contains 54,522 images of 4,102 persons collected from the MSMT17 \cite{MSMT17} database, with each image having a corresponding text description of an average length of 37 words. The vocabulary contains 5554 unique words. Following the data split protocol in \cite{SSAN}, the database is divided into training and testing sets. The training set contains 34674 image-text pairs of 3102 persons, while the testing set consists of 19848 image-text pairs of the remaining 1000 persons. For both databases, we use Rank-K (R@K, K=1, 5, 10, higher is better) to evaluate the performance of different methods.

\subsubsection{Implementation Details}
For images, we use ResNet-50 \cite{resnet} pre-trained on imageNet \cite{imagenet} as the visual backbone with minor modifications: remove the average pooling layer and the fully connected layer, and set the stride of the last convolution layer to 1 for larger feature map. All images are resized to 384$\times$128 before being fed into the image backbone, followed by horizontal flipping for data augmentation. For texts, Bi-LSTM is used as the textual backbone and the text length is unified to $L=100$ before all texts are sent into the textual backbone. The dimension of word embedding is $d_e=768$ for BERT ($d_e=512$ for W2V). The model contains some parameters with dimensions $H=24$, $W=8$, $C=2048$, $d_g=2048$, and $d_c=512$. The scale factors $r_1$, $r_2$, and $r_3$ are set to 32, 256, and 4, respectively. The size of the vocabulary in the training set varies for different databases, with CUHK-PEDES set to $V=5000$ and ICFG-PEDES set to $V=3000$.

During training, we optimize our model using the Adam optimizer with an initial learning rate of 0.001 for the visual backbone and 0.01 for the rest of the network. The learning rate is decayed at the $30$th and $50$th epoch with a decay factor of 0.1. To facilitate training, we utilize the linear warm-up strategy for the first 10 epochs. The model is trained for 70 epochs with a batch size of 64. Both loss margins, $\alpha_1$ and $\alpha_2$, are set to 0.2. The hyper-parameters $\lambda_1$ and $\lambda_2$ are set to 0.1 and 1. Our MANet is implemented in PyTorch and trained on a single RTX3090 24G GPU.
\vspace{-0.25cm}

\begin{table}[!t]\small
\setlength{\abovecaptionskip}{0.1cm}
\centering {\caption{Performance comparison with state-of-the-art methods on ICFG-PEDES. R@1, R@5, and R@10 are listed. '-' denotes that no reported result is available.}\label{Tab:2}
\renewcommand\arraystretch{1.2}
\begin{tabular}{m{2cm}<{\centering}|m{1.5cm}<{\centering}|m{0.9cm}<{\centering}m{0.9cm}<{\centering}m{0.9cm}<{\centering}}
\hline
 \hline
  Methods & Ref & R@1 & R@5 & R@10 \\
  \hline
  CMPM/C~\cite{CMPM} & ECCV18 & 43.51 & 65.44 & 74.26 \\

  SCAN~\cite{SCAN} & ECCV18 & 50.05 & 69.65 & 77.21 \\

  Dual Path~\cite{Dual} & TOMM20 & 38.99 & 59.44 & 68.41 \\

  MIA~\cite{MIA} & TIP20 & 46.49 & 67.14 & 75.18 \\

  ViTAA~\cite{vitaa} & ECCV20 & 50.98 & 68.79 & 75.78 \\

  SSAN~\cite{SSAN} & arXiv21 & 54.23 & 72.63 & 79.53 \\

  TIPCB~\cite{tipcb}   & Neuro22  & 54.96  & 74.72   & \underline{81.89} \\
  
  IVT~\cite{ivt}   & ECCVW22 & 56.04  & 73.60  & 80.22 \\

  SRCF~\cite{SRCF} & ECCV22 & 57.18   & 75.01  & 81.49    \\

  LGUR~\cite{lgur} & MM22  & 57.42 & 74.97  & 81.45    \\

  \hline
  \textbf{MANet-W2V} & - & \underline{57.73} & \underline{75.42} & 81.72 \\
  \textbf{MANet-BERT} & - & \bf{59.44} & \bf{76.80} & \bf{82.75} \\
  \hline\hline
\end{tabular}}
\vspace{-0.4cm}
\end{table}

\subsection{Comparison with State-of-the-art}
In this section, we compare our method with state-of-the-art (SOTA) methods on two public databases. The results on CUHK-PEDES and ICFG-PEDES are shown in Tables~\ref{Tab:1} and ~\ref{Tab:2}, respectively. Our method consistently outperforms the SOTA methods on these two databases.

\begin{table*}[!ht]\small
\setlength{\abovecaptionskip}{0.1cm}
\centering {\caption{Ablation study on different components of our proposed model on CUHK-PEDES.}\label{Tab:3}
\renewcommand\arraystretch{1.2}
\begin{tabular}{m{3.8cm}<{\centering}|m{0.6cm}<{\centering}m{0.6cm}<{\centering}m{0.6cm}<{\centering}m{0.6cm}<{\centering}|m{1cm}<{\centering}m{1cm}<{\centering}m{1.0cm}<{\centering}|m{1.4cm}<{\centering}m{1.4cm}<{\centering}}
\hline
 \hline
  Method  & GA & ILA & RGL & CAF & R@1 & R@5 & R@10 & Params & FLOPs \\
  \hline
  Baseline &  &  &  &  & 57.80 & 77.32 & 84.45 & 66.50M & 8.40 \\

  GA      &\checkmark  &  &  &  & 60.15 & 79.55 & 86.13 & 73.84M & 9.21 \\

  ILA     &  &\checkmark  &  &  & 61.68 & 80.25 & 86.92 & 67.75M & 8.84 \\

  GA+ILA  &\checkmark  &\checkmark  &  &  & 62.43 & 81.14 & 87.35 & 75.09M & 9.65 \\

  GA+ILA+RGL  &\checkmark  &\checkmark  &\checkmark  &  & 63.55 & 81.56 & 87.61 & 82.70M & 11.13 \\

  GA+ILA+CAF  &\checkmark  &\checkmark  &  &\checkmark  & 63.65 & 81.22 & 87.54 & 76.14M & 9.65 \\

  GA+ILA+RGL+CAF (Ours) &\checkmark  &\checkmark  &\checkmark  &\checkmark  & 63.92 & 82.15 & 87.69  & 83.75M & 11.14 \\
  \hline\hline
\end{tabular}}
\vspace{-0.3cm}
\end{table*}

\subsubsection{Results on CUHK-PEDES}
We first evaluate the proposed method on the popular database, \textbf{CUHK-PEDES}, as shown in Table \ref{Tab:1}. For a fair comparison, we classify the methods into two types based on the word embedding ways they are used. Early methods primarily used W2V for word embedding, and the best-performing method SSAN~\cite{SSAN} achieves 61.37\% R@1 accuracy. SSAN explicitly splits an image into multiple parts and uses these parts to guide the generation of part-related local textual features. However, this method relies heavily on the quality of the image parts, which may be ambiguous due to the lack of contextual information. As a result, it is not always reliable to guide the generation of aligned text parts using such image parts. In contrast, our method introduces a set of modality-shared semantic centers to implicitly learn locally aligned features, avoiding hard split and interactions between images and texts, and achieves 63.92\% R@1 accuracy, outperforming SSAN~\cite{SSAN} by 2.55\%. The results highlight the superiority of implicit local feature learning for modality alignment.

With the popularity of Transformer, most subsequent works use the pre-trained BERT model for word embedding. Benefiting from the rich language priors in BERT, such methods achieve superior performance. The typical methods include: (1) SRCF~\cite{SRCF} designs two types of filters to effectively extract the key clues and adaptively align the local features, achieving 64.04\% R@1 accuracy. (2) TIPCB~\cite{tipcb} learns local features with a similar idea to SSAN~\cite{SSAN} and achieves 64.26\% R@1 accuracy. (3) AXM-Net aligns images and texts by designing an AXM-block for complex cross-modal interactions, achieving 64.44\% R@1 accuracy.  (4) LGUR~\cite{lgur} is based on a Transformer decoder to project textual and visual features into a set of query prototypes to learn aligned local features, which is currently the best method (65.25\% R@1 accuracy). Compared with them, our MANet achieves 65.64\% R@1 accuracy and surpasses the existing SOTA method LGUR~\cite{lgur} by 0.39\%, achieving new SOTA performance on CUHK-PEDES without explicit split and cross-modal interactions. Notably, LGUR is a work of the same period as ours, which shares a similar idea to ours in projecting pixel/word features of image/text to a set of shared prototypes for generating local features. This further confirms the superiority of this implicitly locally aligned feature learning. Furthermore, these works~\cite{CMKA, SRCF, lgur} also consider the influence of image background on cross-modal alignment, but ignore the environmental factors which is a crucial factor contributing to the information inequality problem between modalities. Our MANet simultaneously considers these two kinds of noises and tries our best to mitigate their impact on modality alignment.

\subsubsection{Results on ICFG-PEDES}
To prove the generality of the proposed MANet, we also evaluate the performance on another newly released large-scale database, \textbf{ICFG-PEDES}. The comparison results are presented in Table \ref{Tab:2}. Since this database is up-to-date, only a few methods are available for comparison. Our MANet outperforms all existing methods reported on ICFG-PEDES in all metrics under both settings (word embedding via W2V or BERT). Specifically, using W2V for word embeddings, our MANet achieves 57.73\%, 75.42\% and 81.72\% in R@1, R@5 and R@10, respectively. When using BERT for word embeddings, our MANet achieves 59.44\%, 76.80\% and 82.75\% in R@1, R@5 and R@10, respectively, which significantly surpasses the existing SOTA method LGUR~\cite{lgur} by 2.02\%, 1.83\%, 1.30\% on all metrics.








The above results demonstrate the superiority and generality of our MANet for TBPS. This is attributed to its capacity to suppress image-specific information and learn implicitly aligned local features, effectively narrowing the modality gap between images and texts on multiple levels, which will be further demonstrated in the next section.
\vspace{-0.25cm}

\subsection{Ablation Studies}
In this section, we investigate the contribution of each module in the proposed MANet and the effect of important parameters on CUHK-PEDES, including global alignment (GA), implicit local alignment (ILA), image-specific information suppression module (which includes relation-guided localization (RGL) and channel attention filtration (CAF)), and the number $K$ of topic centers. R@1, R@5, and R@10 accuracies (\%) are reported.

\subsubsection{Effectiveness of Components}
To clearly illustrate the effectiveness of each component, we incrementally add them to the model and report the experimental results in Table \ref{Tab:3}. Following SSAN \cite{SSAN}, Baseline refers to a basic global alignment, which means to perform max-pooling on image and text feature maps ($\bm F$ and $\bm E$), followed by embedding image and text features into a common space for global alignment through a modality-shared 1×1 convolutional layer. We train Baseline using the same training strategy as ours. The first row shows the results of Baseline.

GA represents replacing the global alignment in Baseline as our global alignment module, the result is shown in the second line. Compared with Baseline, GA improves R@1, R@5, and R@10 by 2.35\%, 2.23\%, and 1.68\% respectively. When the ILA module is added to Baseline, R@1, R@5, and R@10 accuracies are significantly improved by 3.88\%, 2.93\%, and 2.47\% over Baseline, which demonstrates that this method of implicitly learning locally aligned features by a set of modality-shared semantic topic centers can effectively model local fine-grained correspondence between images and texts, and narrow the semantic gap across modalities. The joint deployment of GA and ILA can further improve performance, with R@1 accuracy on CUHK-PEDES achieving 62.43\%, which has surpassed all methods of the same type in Table \ref{Tab:1}, further proving the superiority of our method. Moreover, the result also demonstrates that GA and ILA can work cooperatively with each other to align images and texts from different granularities.

Although the combination of GA and ILA has achieved superior performance, RGL and CAF can further improve R@1 accuracy by 1.12\% and 1.22\% respectively based on the above results, since RGL and CAF consider the influence of image background and environmental factors, and alleviates the information inequality problem between modalities. The joint deployment of RGL and CAF can further improve performance, and achieve the best. All the above results show that each module in the proposed MANet can play a positive role effectively in alleviating the modality gap.

\subsubsection{Effectiveness of Image-Specific Information Suppression}
The attention mechanism is crucial in the ISS module. To prove the advantages of RGL and CAF in suppressing image-specific information, we conduct a series of comparative experiments, including directly using instance normalization (IN) to eliminate environmental factors, removing the identity-relevant content consistency loss $\mathcal{L}_{cons}$ designed in CAF, and replacing RGL with existing attention methods such as CBAM \cite{cbam}, Non-Local (NL) \cite{non-local}, RGA \cite{RAG}. A$\rightarrow$B means replacing A with B, leaving the rest unchanged. The results are presented in Table \ref{Tab:4}.

When replacing CAF with IN, there is a drop in R@1 by 1.66\% over MANet as loss of some discriminative identity-relevant information along with the elimination of environmental factors. CAF overcomes this problem by using channel attention to restore useful identity information from the IN-removed information. Furthermore, CAF incorporates an identity-relevant content consistency loss to promote information restitution and enhance feature discrimination, resulting in an improvement of nearly 0.5\% in R@1 accuracy. In addition, we compare RGL with some existing attention methods, including (1) CBAM~\cite{cbam}, which directly feeds feature maps into a standard convolutional layer to generate weights for each spatial location. (2) Non-Local (NL)~\cite{non-local}, which strengthens the features of the target position by computing a weighted summation of features of all positions (sources), where the weights are the pairwise relations between the target position and all other positions. (3) RAG~\cite{RAG}, which models the global structural relations with other locations for each location, and then infers the attention of the current location based on the feature of the current location and corresponding structural relations.

\begin{table}[!t]\small
\setlength{\abovecaptionskip}{0.1cm}
\centering {\caption{Ablation study of image-specific information suppression module on CUHK-PEDES.}\label{Tab:4}
\renewcommand\arraystretch{1.2}
\begin{tabular}{m{2.6cm}<{\centering}|m{0.7cm}<{\centering}m{0.7cm}<{\centering}m{0.7cm}<{\centering}|m{0.8cm}<{\centering}m{0.8cm}<{\centering}}
\hline
 \hline
  Method  & R@1 & R@5 & R@10 & Params & FLOPs\\
  \hline
  MANet            & 63.92 & 82.15 & 87.69 & 83.75M & 11.14  \\
  \hline
  CAF$\rightarrow$IN      & 62.26   & 80.54   & 87.02 & 82.71M & 11.13 \\

  CAF (w/o $\mathcal{L}_{cons}$)  & 63.47  & 81.38  & 87.91 & 83.75M & 11.14\\

  RGL$\rightarrow$CBAM \cite{cbam}  & 63.03   & 81.11   & 87.90 & 76.67M & 9.66\\

  RGL$\rightarrow$NL \cite{non-local}  & 62.09   & 81.06   & 87.61 & 84.54M & 11.27 \\

  RGL$\rightarrow$RAG \cite{RAG}   & 63.17   & 81.47   & 87.56 & 77.73M & 9.96 \\
  \hline\hline
\end{tabular}}
\vspace{-0.5cm}
\end{table}

The results indicate a certain level of degradation in performance. As shown in Figure~\ref{Fig:1}, the image regions that correspond to textual descriptions are structured yet irregular. It is not reasonable for CBAM~\cite{cbam} to use a regular local context, i.e., a convolution operation of filter size 7×7, to determine the attention of a feature position. While Non-Local~\cite{non-local} introduces global relations to strengthen the features of the target position by aggregating information from all positions, Cao~\emph{et al}.~\cite{Cao} have shown that the aggregation weights are invariant to the target position and only locally determined by source feature itself, which does not fully utilize the global structural information. RAG~\cite{RAG} overcomes these shortcomings and achieves better performance. In RAG, the measurement of the relation between features is crucial. RAG obtains a relation scalar by calculating the inner product between feature vectors, which only reflects the degree of correlation between features and lacks detailed correspondence. In contrast, our RGL computes a relation vector by the subtraction between feature vectors, which results in a 0.75\% improvement in R@1 over RAG. Guided by these detailed relations, the network can better understand images and highlight modality-shared body regions to better align with texts under the supervision of the cross-modal ranking loss.

\subsubsection{Different Aggregation Strategy}
In the proposed implicit local alignment (ILA) module, image and text features are aligned by adaptively aggregating them into a set of shared semantic topic centers. The feature aggregation strategy plays a crucial role in this process. We evaluated different aggregation strategies and present the results in Table \ref{Tab:5}. NetVLAD \cite{netvlad} is a classic feature aggregation method that generates a set of local features by computing the weighted sum of the residuals between each pixel (word) feature and the centers. However, it introduces too much uncertainty, leading to severe performance degradation (2.83\% drop over our ILA). In contrast, we compute the relations of all pixel (word) features to the center as weights, and the weighted sum of all pixel features to generate the local feature for the center, rather than residuals. LGUR~\cite{lgur} implements similar ideas to ours based on Transformer's decoder. For a fair comparison, we replace our ILA with the decoder in LGUR. The main difference between ILA and Decoder is the weight calculation method when aggregating features. Decoder generates a weight scalar by calculating the inner product between the feature and the center, and treats each channel of the feature equally in the aggregation feature. While ILA generates a weight vector with detailed correspondence by computing the subtraction between the feature and the center, and considers both channel and spatial dimensions when aggregating features. As a result, our method performs better. However, Decoder benefits from multi-head attention (MHA), which is equivalent to dividing the feature channel into multiple parts and calculating a weight scalar for each part. Compared with directly calculating the inner product between the entire vectors, MHA provides slightly more detailed correspondence, resulting in only a slightly lower performance than our ILA. However, MHA also brings higher computational costs. When there is no multi-head (H=1), its performance is much lower than our ILA. ILA (IP) means that the assignment to each center is calculated by inner product (IP) operation, which shows a serious performance drop. These results all demonstrate that a more detailed correspondence can provide more accurate assignments when aggregating features to each semantic center.

\begin{table}[!t]\small
\setlength{\abovecaptionskip}{0.1cm}
\centering {\caption{Effects of different feature aggregation on CUHK-PEDES. 'IP' represents an inner-product operation.}\label{Tab:5}
\renewcommand\arraystretch{1.2}
\begin{tabular}{m{2.6cm}<{\centering}|m{0.65cm}<{\centering}m{0.65cm}<{\centering}m{0.65cm}<{\centering}|m{0.9cm}<{\centering}m{0.8cm}<{\centering}}
\hline
 \hline
  Method  & R@1 & R@5 & R@10 & Params & FLOPs \\
  \hline
  NetVLAD \cite{netvlad}  & 61.09   & 80.99  & 86.81 & 82.52M & 10.69 \\

  Decoder~\cite{lgur} & 63.63 & 81.86 & 87.83  & 126.57M & 15.86 \\

  Decoder (H=1)~\cite{lgur} & 61.98 & 80.19 & 87.15  & 101.41M & 12.03 \\

  ILA (IP)  & 61.19   & 80.04   & 87.04 & 83.55M & 10.99\\

  ILA (Ours) & 63.92 & 82.15 & 87.69 & 83.75M & 11.14 \\
  \hline\hline
\end{tabular}}
\vspace{-0.2cm}
\end{table}

\begin{table}[!t]\small
\setlength{\abovecaptionskip}{0.1cm}
\centering {\caption{Effects of center initialization schemes of ILA on CUHK-PEDES.}\label{Tab:6}
\renewcommand\arraystretch{1.2}
\begin{tabular}{m{4.6cm}<{\centering}|m{0.8cm}<{\centering}m{0.8cm}<{\centering}m{0.8cm}<{\centering}}
\hline
 \hline
  Method  & R@1 & R@5 & R@10 \\
  \hline
  Standard normal distribution (Ours)	  & 63.92	& 82.15	& 87.69  \\
  Uniform distribution	              & 63.71	& 81.89	& 87.96  \\
  Kaiming\_normal distribution	      & 63.78	& 81.94	& 88.08  \\
  Xavier\_normal distribution	          & 63.43	& 81.56	& 87.91  \\
  All-zero matrix (1e-8)	              & 63.65	& 81.99	& 87.93  \\
  All-one	matrix                        & 63.61   & 82.08	& 88.01  \\
  Identity matrix                       & 63.91	& 81.86	& 87.72  \\
  Constant matrix (0.3)	              & 63.63	& 81.68	& 87.67  \\
  Orthogonal matrix	                  & 63.89   & 81.92 & 88.00  \\
  \hline\hline
\end{tabular}}
\vspace{-0.2cm}
\end{table} 

\subsubsection{Center Initialization}
In our experiments, we randomly initialize the semantic centers $\{\bm c_1, \bm c_2, ..., \bm c_K\}\in \mathbb{R}^{K\times d_c}$ in the ILA module using a standard normal distribution. To evaluate the impact of center initialization on performance, we also consider several common initialization schemes such as uniform distribution initialization, Kaiming\_normal distribution initialization, Xavier\_normal distribution initialization, all-zero initialization, all-one initialization, identity matrix initialization, constant initialization, and orthogonal initialization. Table~\ref{Tab:6} summarizes the results, showing that our method achieves superior performance under different initialization schemes. This demonstrates the robustness of our method to center initialization. While reasonable weight initialization can accelerate model convergence and facilitate global optimal solution search, especially for models with a large number of parameters, our center is only a simple parameter matrix with limited dependence on initialization.

\subsubsection{Number of Topic Centers}
\begin{figure}[!t]
\centering
\setlength{\abovecaptionskip}{-4pt}
\includegraphics[width=3.2in,height=1.8in]{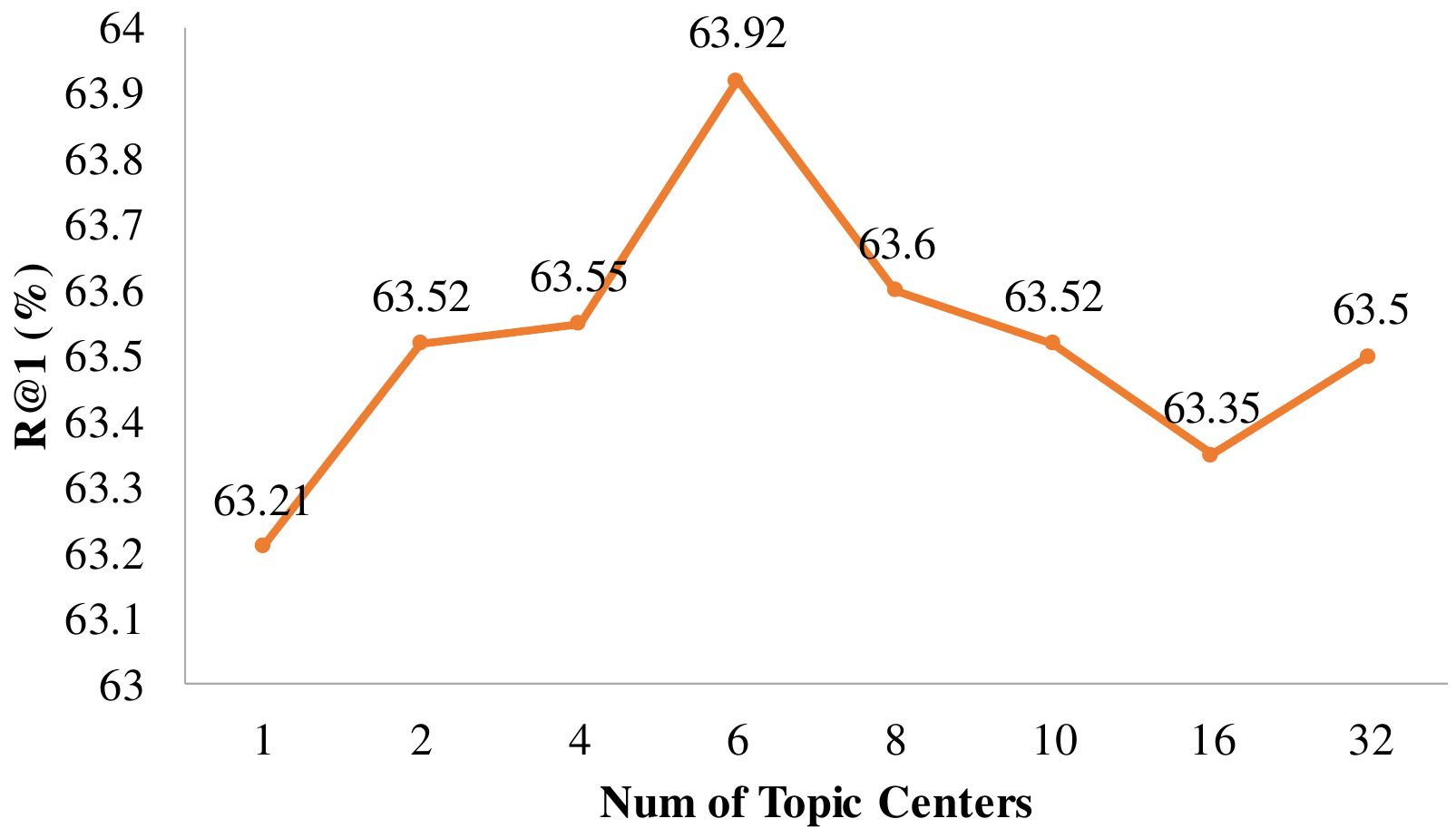}\\
\caption{Effect of different number of semantic topic centers.}
\vspace{-0.2cm}
\label{Fig:5}
\end{figure}

The value of $K$ determines the granularity of the local feature, which is crucial for establishing fine-grained correspondence between modalities. We select the most appropriate value of $K$ from the range of 1 to 32, as illustrated in Figure \ref{Fig:5}. The results indicate that if $K$ is too small, the granularity of the obtained local features is insufficient, and establishing fine-grained correspondence becomes difficult. When $K$=1, the learned features represent a global one. As $K$ increases, the R@1 accuracy gradually improves but begins to decrease after it is greater than 6. We believe that too large granularity will lead to semantic fragmentation and severely destroy the semantics of images and texts. When $K$=6, our proposed method achieves the best results.

\subsubsection{Cost Complexity}
Table~\ref{Tab:3} presents the cost complexity of our method, reporting the number of model parameters (Params) and the number of floating-point operations for an input image (FLOPs). Compared with Baseline, our MANet incurs higher computational cost primarily due to the GA (fully connected layer) and RGL modules (attention mechanism), while the CAF and ILA modules, which are the main contributions of our paper, require very low Params and FLOPs, leading to significant performance improvement at only a very low computational cost. Furthermore, Table~\ref{Tab:7} compares the inference times (across all test queries) of different methods. Both Baseline and GA are simple global alignments. On this basis, we introduce the local branch, but just one more local vector (by concatenating multiple local features) for similarity calculation, so only a small increase in inference time.

\begin{figure}[!t]
\centering
\setlength{\abovecaptionskip}{-2pt}
\includegraphics[width=3.2in, height=1.6in]{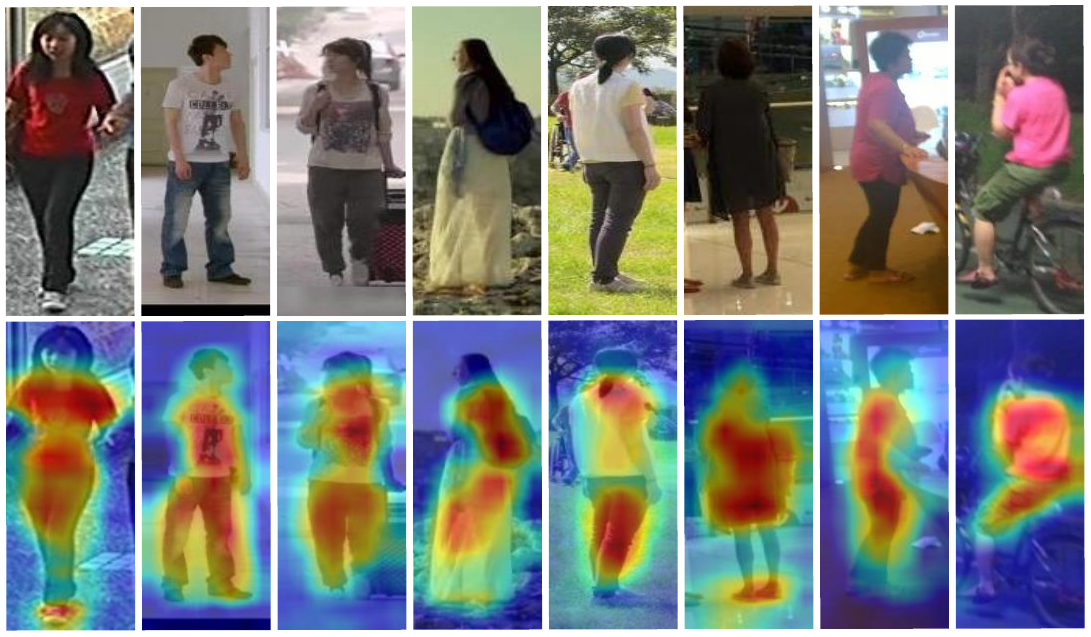}\\
\caption{Visualization of image response maps in the RGL module.}
\vspace{-0.15cm}
\label{Fig:6}
\end{figure}

\begin{figure}[!t]
\centering
\setlength{\abovecaptionskip}{-2pt}
\includegraphics[width=3.2in, height=3.2in]{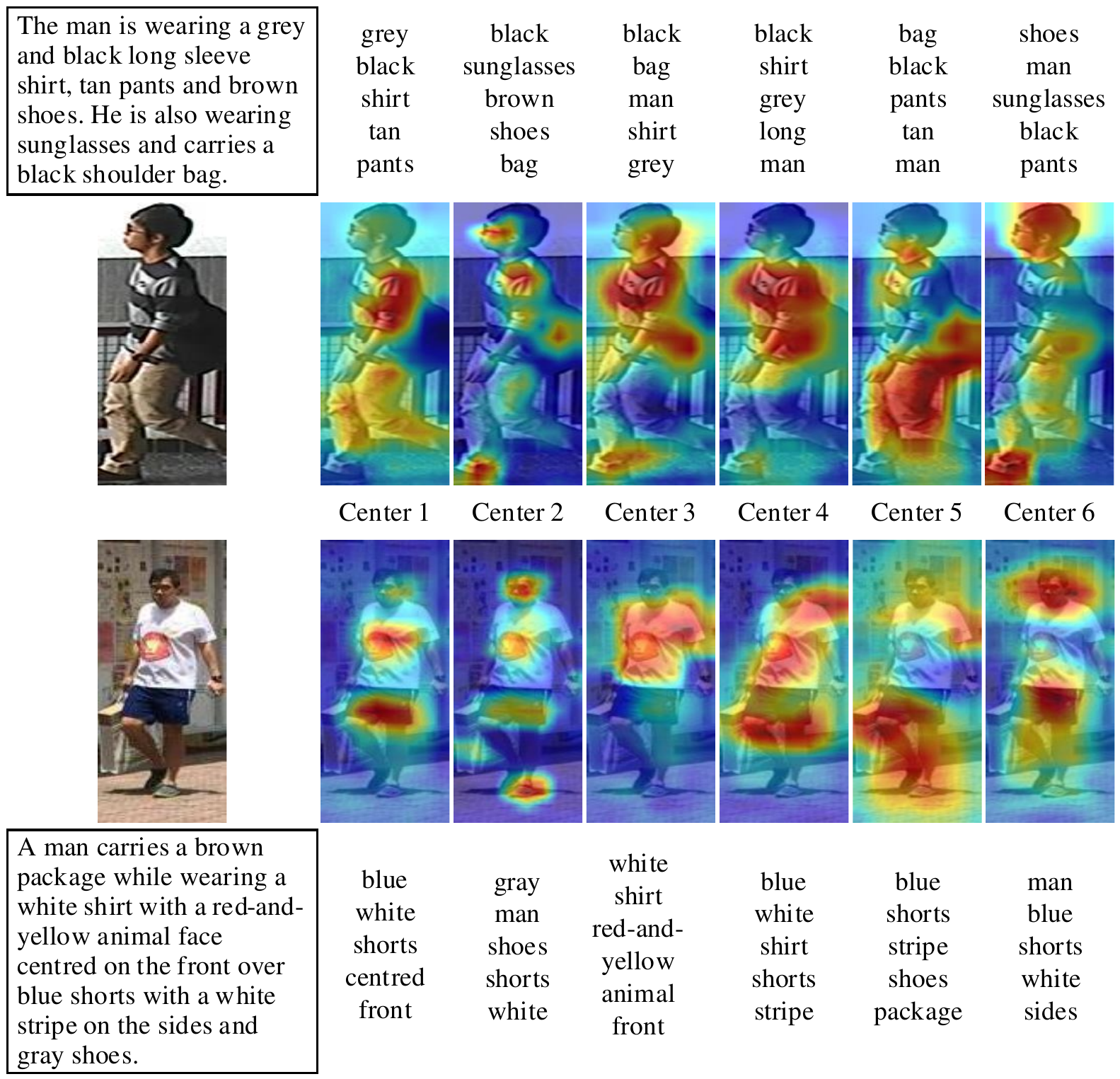}\\
\caption{Visualization of image response maps and top-5 response words for each center in the ILA module.}
\vspace{-0.15cm}
\label{Fig:7}
\end{figure}

\begin{table}[!t]\small
\setlength{\abovecaptionskip}{0.1cm}
\centering {\caption{Comparison of inference time (s) on CUHK-PEDES.}\label{Tab:7}
\renewcommand\arraystretch{1.2}
\begin{tabular}{m{1.6cm}<{\centering}|m{0.8cm}<{\centering}m{0.8cm}<{\centering}m{2.0cm}<{\centering}|m{0.7cm}<{\centering}}
\hline
 \hline
  Method    & Params  & FLOPs & Inference Time & R@1\\
  \hline
  Baseline  & 66.50M & 8.40  & 13.722s & 57.80 \\
  GA        & 73.84M & 9.21  & 14.175s & 60.15 \\
  MANet     & 83.75M & 11.14 & 15.422s & 63.92 \\
  \hline\hline
\end{tabular}}
\vspace{-0.2cm}
\end{table}

\subsubsection{Qualitative Results}
Figure~\ref{Fig:6} visualizes some typical image response maps generated by the RGL module, which demonstrates that this module effectively highlights human body regions in images while suppressing background regions. Additionally, Figure~\ref{Fig:7} showcases the visualization results of image response maps and top-5 response words for each center in the ILA module. From these results, we can draw the following conclusions: (1) Each center in the ILA module is capable of adaptively mining image regions that contain complete contextual information corresponding to texts, thus establishing a fine-grained correspondence between images and texts. (2) The combination of features corresponding to these centers can fully express pedestrians at a fine level.

\begin{figure}[!t]
  \centering
  \setlength{\abovecaptionskip}{-2pt}
  \includegraphics[width=3.2in,height=3.0in]{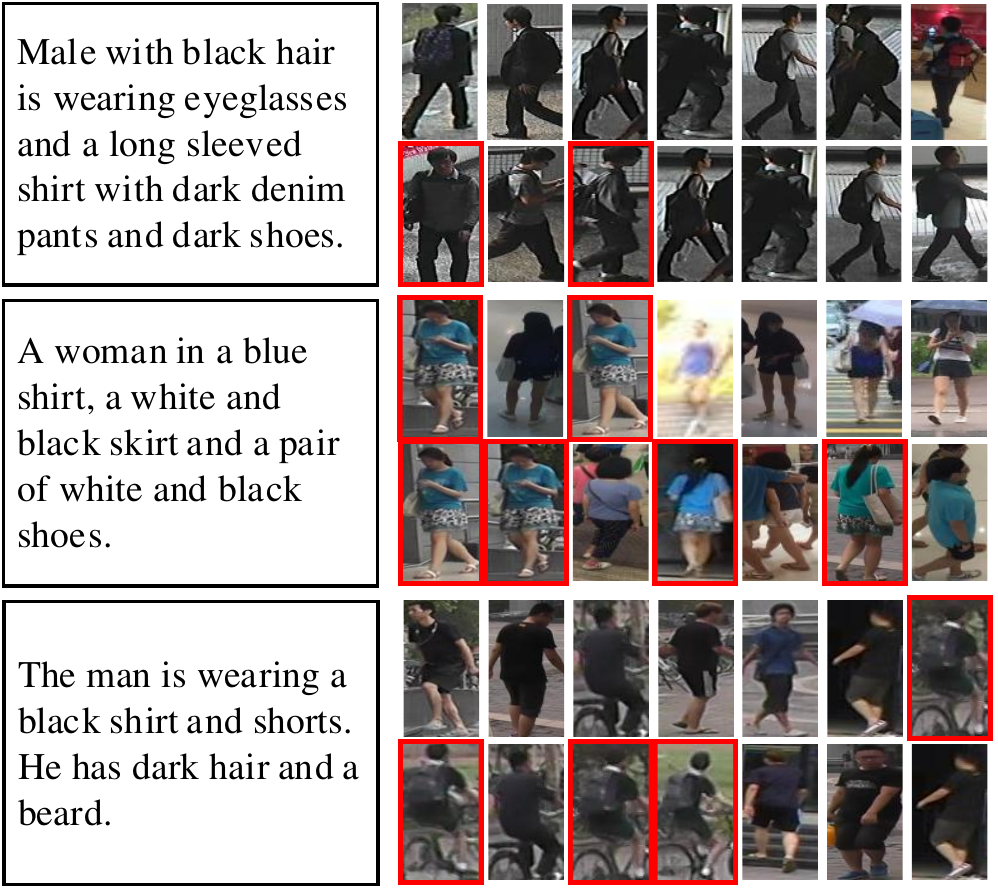}\\
  \caption{Comparison of top-7 retrieval results on CUHK-PEDES between Baseline (the 1st row) and our MANet (the 2nd row) for the same query texts. The correct results are marked by red rectangles.}
  \vspace{-0.5cm}
  \label{Fig:8}
\end{figure}

Figure \ref{Fig:8} illustrates the top-7 retrieval results of our method and Baseline for a given text query. As can be seen from the figure, our method achieves more accurate retrieval results. In some cases where Baseline fails, our method can correctly identify the desired image within the top-3 results. Although there are instances where the top-7 retrieved images contain incorrect results, the semantic attributes described in the text query are present in almost all of the retrieval results. This demonstrates that our method effectively aligns the modalities of images and texts and can accurately establish detailed correspondences between them. Furthermore, our results indicate that color information plays a dominant role in text-to-image retrieval. However, as mentioned earlier, environmental factors can cause image distortion and lead to the loss of semantic information. Therefore, relying solely on color information as the leading retrieval clue is not always reliable. In our future work, we aim to mine deeper semantic information to improve the accuracy and robustness of existing methods.

\section{Conclusion}
In this paper, we introduce MANet, an efficient joint multi-level alignment network designed to alleviate the modality gap between images and texts for TBPS. Our method aims to implicitly learn the local fine-grained correspondence between images and texts while addressing the information inequality problem between modalities. Specifically, MANet leverages an Image-Specific Information Suppression module to alleviate the information inequality problem. This module includes a Relation-Guided Localization submodule that models global relation information to help localize the human body region, as well as a Channel Attention Filtration submodule that applies instance normalization and channel attention to eliminate environmental factors while preserving identity information. Next, in the Implicit Local Alignment module, image and text features are adaptively aggregated into a set of shared semantic topic centers, which enables the implicit learning of locally aligned features without extra supervision and cross-modal interactions. Finally, a Global Alignment module is introduced to provide a global cross-modal measurement that complements the local perspective. The proposed MANet is optimized in an end-to-end manner. We demonstrate the superiority and effectiveness of MANet through both qualitative and quantitative experimental results, as well as extensive ablation studies.


%





\ifCLASSOPTIONcaptionsoff
  \newpage
\fi



%
\bibliography{mybibfile}

\end{document}